\documentclass[conference]{IEEEtran}
\IEEEoverridecommandlockouts

\usepackage{amsmath,amssymb,amsfonts}
\usepackage{algorithmic}
\usepackage{graphicx}
\usepackage{textcomp}
\usepackage{xcolor}
\usepackage{siunitx}
\usepackage[numbers]{natbib}
\usepackage{hyperref}
\def\BibTeX{{\rm B\kern-.05em{\sc i\kern-.025em b}\kern-.08em
    T\kern-.1667em\lower.7ex\hbox{E}\kern-.125emX}}

\DeclareMathOperator*{\argmax}{arg\,max}

\begin{document}

\title{Putting the Iterative Training of Decision Trees to the Test on a Real-World Robotic Task

\thanks{Author R.C.E.'s work was partially supported by the research training group “Dataninja” (Trustworthy AI for Seamless Problem Solving: Next Generation Intelligence Joins Robust Data Analysis) funded by the German federal state of North Rhine-Westphalia}
}

\author{\IEEEauthorblockN{Raphael C. Engelhardt}
\IEEEauthorblockA{\textit{
Cologne Institute of Computer Science} \\
\textit{TH Köln} \\
Gummersbach, Germany \\
0000-0003-1463-2706}
\and
\IEEEauthorblockN{Marcel J. Meinen}
\IEEEauthorblockA{\textit{
Institute of General Mechanical Engineering} \\
\textit{TH Köln} \\
Gummersbach, Germany \\
0009-0006-6970-522X}
\and
\IEEEauthorblockN{Moritz Lange}
\IEEEauthorblockA{\textit{
Institute for Neural Computation} \\
\textit{Ruhr-University Bochum}\\
Bochum, Germany \\
0000-0001-7109-7813}
\and
\IEEEauthorblockN{Laurenz Wiskott}
\IEEEauthorblockA{\textit{
Institute for Neural Computation} \\
\textit{Ruhr-University Bochum}\\
Bochum, Germany \\
0000-0001-6237-740X}
\and
\IEEEauthorblockN{Wolfgang Konen}
\IEEEauthorblockA{\textit{
Cologne Institute of Computer Science} \\
\textit{TH Köln} \\
Gummersbach, Germany \\
0000-0002-1343-4209}

}

\maketitle

\begin{abstract}
In previous research, we developed methods to train decision trees (DT) as agents for reinforcement learning tasks, based on deep reinforcement learning (DRL) networks. The samples from which the DTs are built, use the environment's state as features and the corresponding action as label. To solve the nontrivial task of selecting samples, which on one hand reflect the DRL agent's capabilities of choosing the right action but on the other hand also cover enough state space to generalize well, we developed an algorithm to iteratively train DTs.

In this short paper, we apply this algorithm to a real-world implementation of a robotic task for the first time. 
Real-world tasks pose additional challenges compared to simulations, such as noise and delays.
The task consists of a physical pendulum attached to a cart, which moves on a linear track. By movements to the left and to the right, the pendulum is to be swung in the upright position and balanced in the unstable equilibrium. Our results demonstrate the applicability of the algorithm to real-world tasks by generating a DT whose performance matches the performance of the DRL agent, while consisting of fewer parameters. This research could be a starting point for distilling DTs from DRL agents to obtain transparent, lightweight models for real-world reinforcement learning tasks.
\end{abstract}


\section{Introduction}
In recent studies, we explored the possibilities of deriving decision trees (DTs) from trained deep reinforcement learning (DRL) agents. Using samples, consisting of the states of the environment as features and the corresponding actions as labels, the reinforcement learning (RL) problem can be translated to a supervised learning problem~\cite{lod2022}. The choice of samples proved to be of determining importance for the successful training of the DTs. Using episodes of well-performing DRL agents, only a very narrow region of the state space is covered, sometimes too small for a DT to represent a successful policy. We developed an algorithm that iteratively uses DTs to explore regions of the state space during episodes and 
the DRL agent as a sort of ``teacher'' to label these states with the ``correct'' actions~\cite{engelhardt2023}. We successfully tested this algorithm on a variety of classical RL challenges and were able to find oblique DTs\footnote{The decision nodes of oblique DTs compare a linear combination of the features to a threshold. Oblique DTs therefore use oblique hyperplanes to partition the data.} that not only match the performance of their DRL ``teachers'' but in most cases even surpass them.

The experiments had, however, only been conducted on simulated control problems from the gymnasium suite~\cite{gymnasium}. In this short paper, we explain how we applied the iterative algorithm to a real-world robotic task. We describe the implementation of the task and the experimental setup, and show how our results prove the applicability of the algorithm. We also discuss limitations and potential improvements for further experiments.

\section{Related Work}
The interplay of RL and robotic applications in the real world has been the subject of interest in various studies.

\citeauthor{quadrupedal}~\cite{quadrupedal} apply RL to train the locomotion of a blind quadrupedal robot in simulation and test its robustness in the real world on rough terrain, unseen during training. This study is relevant, as it shows the feasibility of transferring a policy from the simulated to the real domain in a robotics task,  especially since the complexity of reality is expensive to approximate in simulation.

\citeauthor{droneracing}~\cite{droneracing} apply DRL to find near-time-optimal tracks for drone racing. The challenge consists of flying a quadrotor drone along a predefined track consisting of various gates the drone must fly through in the shortest amount of time possible. They rely on the DRL algorithm PPO to find a policy mapping the quadrotor's state and observations about future gates to thrust of the four individual rotors. Although most of their experiments are conducted in simulation, in Section V-F, they briefly deploy a generated trajectory to a real-world flight. While \citeauthor{droneracing} report high speeds of the drone, they also encounter large tracking errors and acknowledge the need for further research on the topic.

Another important robotic task is the grasping of objects. \citeauthor{grasp}~\cite{grasp} train a CNN to predict the likelihood of a successful grasp based on pixel data from a camera and apply a servoing mechanism to issue the robot's motors accordingly. To produce the $800\,000$ grasp attempts for training, a cluster of $6-14$ robots was used. This not only parallelizes the collective experience-making, but the authors also state that the subtle differences of the individual setups contributed to the rich dataset, allowing for a successful training of hand-eye coordination. Notably, while the first two publications mentioned above evaluate models trained in simulation when applied to the real domain, in this third one---like ours---already the training is conducted in the real-world setting.

An interesting study on the combination of robotics, DTs, and RL was published by \citeauthor{nao}~\cite{nao}. Contrary to our approach, they did not use DTs to approximate the DRL agent's policy but rather to approximate the environment. In model-based RL, the transition and reward functions are learned, to then simulate possible actions inside the found model. One of the challenges is the large amount of exploration needed to find an accurate model of the domain. The authors leverage the generalization capabilities of DTs to efficiently learn the transition and reward functions. They successfully evaluated their technique on the task of the Aldebaran Nao humanoid robot performing a penalty kick in the RoboCup robot soccer competition, not only in simulation but also in the real-world setting.

For a more complete overview on the challenges of real-world robotics in conjunction with RL and current studies on this subject, we refer the reader to the survey by \citeauthor{survey}~\cite{survey}.

\section{Robotic RL Task}
The robotic task studied in this paper is a real-world implementation of the CartPole Swing-up (CPSU) challenge. An inverted physical pendulum is attached to a cart with an unactuated hinge. The cart itself can move freely on a linear track. By pushing the cart to the left or to the right, the inverted pendulum is to be swung up to the point of unstable equilibrium. Once this first goal has been achieved, the pendulum must be balanced in the upright position for the rest of the episode. 
The pole balancing problem (without the swing-up phase), described in \cite{cartpole}, is a popular environment for RL and part of the widely-used benchmark suite gymnasium \cite{gymnasium}. A simulation of the full problem, including the swing-up phase, is given by \cite{freeman}.

\begin{figure}
    \centering
    \includegraphics[width=0.8\linewidth]{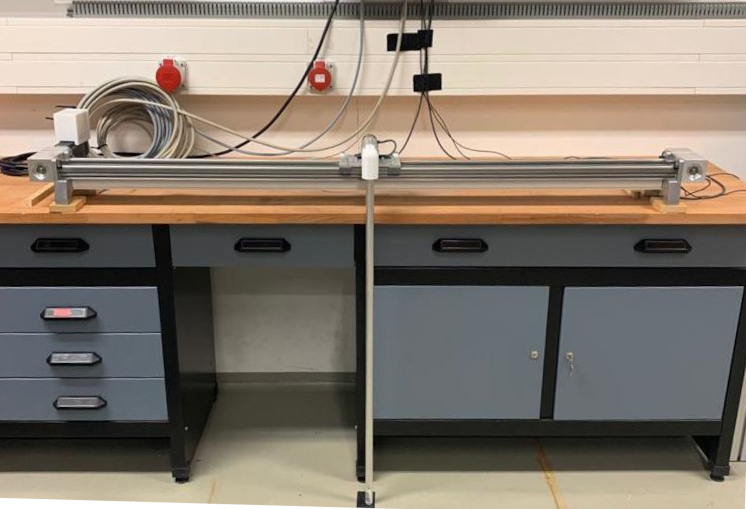}
    \caption{Photograph of the real-world implementation of the CPSU task. Adapted from~\cite{nayante}}
    \label{fig:setup}
\end{figure}

In the real-world implementation in the Lab for Applied Artificial Intelligence at TH Köln -- University of Applied Sciences, the physical pendulum is a \SI{975}{\milli \metre} aluminum rod attached to a cart. 
The cart can be moved on a \SI{1500}{\milli \metre} track by a belt actuated by a DC servo motor (by software, the movement is restricted to \SI{\pm390}{\milli \metre}). The motor is controlled via a Raspberry Pi mini-computer and an STM microcontroller. The setup can be seen in Figure~\ref{fig:setup}.

Sensors measure the four observables of the RL problem:
\begin{itemize}
    \item Angle of the rod $u$ in the range $[-180, 180]$ normalized for the RL agent to $u' \in [-1, 1]$. The coordinate system was chosen such that the pendulum hanging down in the stable equilibrium corresponds to $0$, while the upright unstable equilibrium is defined as $180$.
    \item The rod's angular velocity $\dot{u}$
    \item Cart's position $y$ normalized via division by $390$. The center of the track corresponds to $y=0$.
    \item The cart's linear velocity $\dot{y}$
\end{itemize}

At each timestep, the RL agent can choose to move the cart to the right, to the left, or not at all. It is therefore an RL problem with a discrete action space. The selected action will then be executed by the motor for a duration of \SI{0.1}{\second}, before the next time step begins. The episode terminates after $1000$ timesteps, if the cart leaves the boundaries of the linear track $|y| > 390$, or if the angular velocity exceeds a safe range $|\dot{u}| > 100$. 
After each episode, a cool-down phase of \SI{90}{\second} ensures the pendulum comes to a complete standstill so that the next episode starts in the same condition.

The reward function is chosen such as to reward the agent for keeping the rod in the upright position and the cart preferably centered: 

\begin{equation}\label{eq:reward}
r = \underbrace{\frac{1}{2} \cdot \left(1-\cos\left(\frac{u\cdot \pi}{180}\right)\right)}_{\text{reward angle}} \cdot \underbrace{\cos \left(\frac{\pi}{2} \cdot \frac{y}{390}\right)}_{\text{reward position}} + r_{\mathrm{bonus}}
\end{equation}
where
\begin{equation}\label{eq:rewardbonus}
	r_{\mathrm{bonus}} = \begin{cases} 
			10 & \text{if } |u| > \frac{175}{180} \text{ and } |\dot{u}| < \frac{6}{40} \\
			0 & \text{otherwise.}
		\end{cases}
\end{equation}

This contains a bonus reward $r_{\mathrm{bonus}}$, issued for each time step in which the pendulum is considered close enough to the zenith, defined as sufficiently large angle $u$ and small angular velocity $\dot{u}$ in the first case of Equation~\eqref{eq:rewardbonus}.

\section{Experimental Setup}
The iterative algorithm was developed to derive DTs from trained DRL agents. Here, we only outline its main working principles and refer to~\cite{engelhardt2023} for an in-depth discussion. Initially, a number $N_b$ of base samples are collected during evaluation episodes of the DRL agent. The samples consist of the states of the environment and the executed actions of the DRL agent. A number $N_T$ of DTs of depth $d$ is trained on these samples (iteration $0$). The performance of these DTs is then evaluated in a number $n_e$ of evaluation episodes. Next, the states of the best-performing tree are labeled with the actions suggested by the DRL agent's policy.
These new state-action-pairs are added to the previous samples and a new set of DTs is trained on this new, enriched, dataset (iteration $1$). This interplay of using the exploration of DTs to generate states and the well-performing policy of the DRL agent to provide the corresponding actions continues until a stopping criterion is met.

For our experiment, we trained $N_T=10$ oblique DTs of depth $d=10$ at each iteration using the algorithm and implementation of~\cite{opct} (OPCT). This best-out-of-$10$-approach is necessary given the rather high fluctuations of different DTs trained on identical data due to initialization and has already been applied for experiments on simulated environments in~\cite{engelhardt2023}. Due to the higher time costs of real-world evaluation episodes compared to simulated ones, we only use $n_e=5$ evaluation episodes for each DT in each iteration. The experiment comprised ten iterations in total, as did the experiments on simulated environments in~\cite{engelhardt2023}.

The DRL agent we use to label the states and whose behavior we want to approximate with DTs was trained on the same system as part of a different research project~\cite{nayante}.
The DQN consists of two dense layers with $64$ neurons each and uses the $\tanh$ activation function. It comprises a total of $4675$ trainable parameters. 
In $100$ evaluation episodes, the DQN agent reaches an average return of $\overline{R} = 7138.83 \pm 1517.47$ on the real-world pendulum, showing that the oracle itself is quite good, but not perfect and with non-negligible fluctuations in the performance.

It should be noted, that the real-world implementation entails additional challenges for the DRL agent compared to the simulated environment. These include a time delay between issuing a prediction and actually executing the corresponding action, with consequences for the successful training of RL agents that are hard to anticipate. Additionally, sensor noise affects the agent's perception of the system's state, while other physical factors, such as wear and mechanical play, introduce further challenges.

\section{Results}
We present the results of the experiments. First, we show the exploratory investigation of the base samples followed by the results of the iterative DT training.

\subsection{Base Samples}
\begin{figure}
    \centering
    \includegraphics[width=\linewidth]{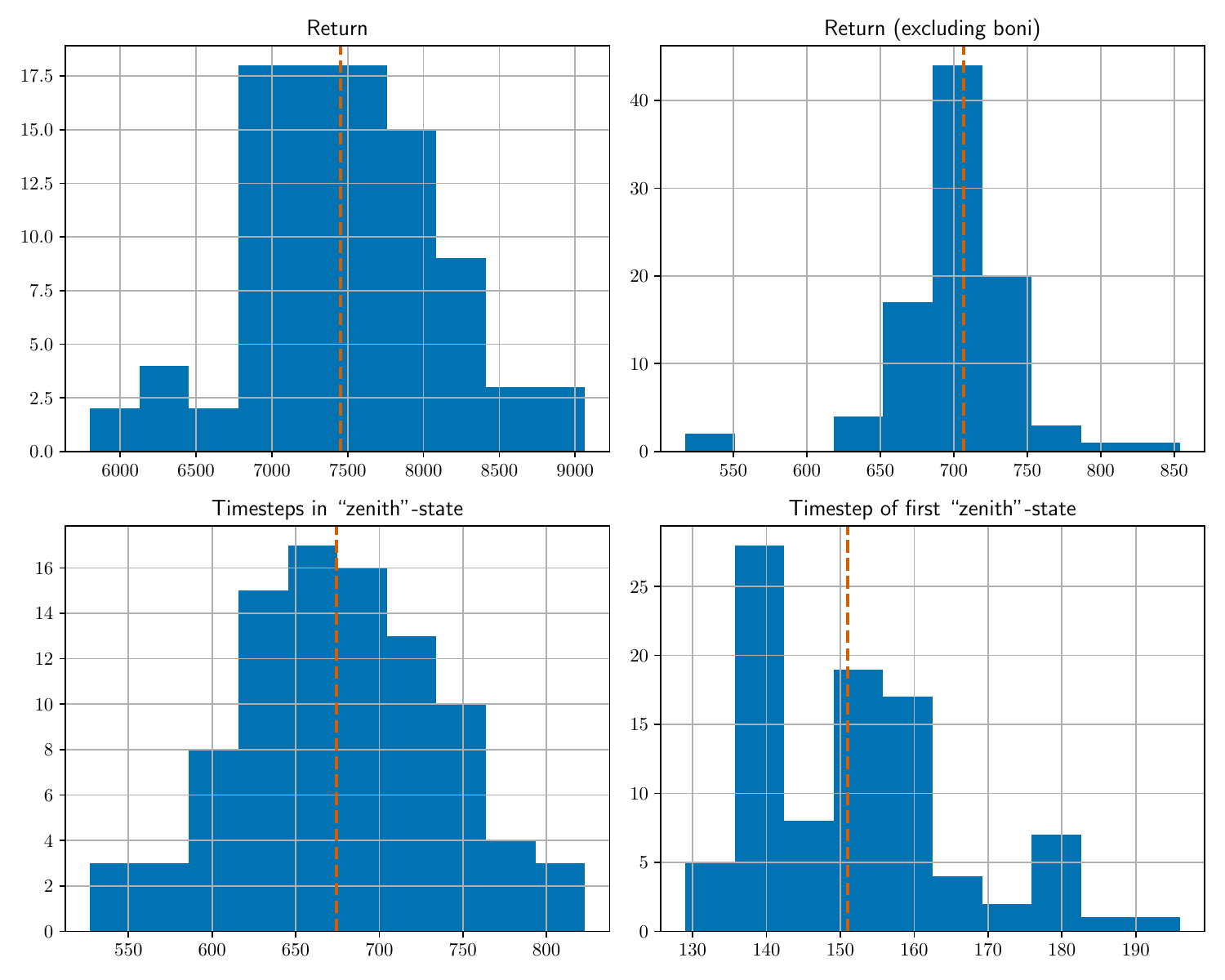}
    \caption{Histograms showing the performance of the $92$ episodes of the DQN agent. The dashed red vertical line marks the median.}
    \label{fig:stat_base_samples}
\end{figure}

The base samples were collected during $100$ episodes of the DNQ agent. These episodes were filtered according to two criteria to ensure high-quality data. First, we discarded the episodes, in which the pendulum never reached the state defined as zenith (see Equation~\eqref{eq:rewardbonus}). Two episodes were excluded as a consequence. Additionally, we filtered out six additional outlier episodes (i.e., episodes whose return was not within $[\mathrm{Q}1 - 1.5 \cdot \mathrm{IQR}, \mathrm{Q}3 + 1.5 \cdot \mathrm{IQR}]$). In the remaining $92$ episodes, the median time for the pendulum to reach the zenith-state for the first time was $151$ timesteps and the average return was $\overline{R} = 7468.72 \pm 621.72$.
Figure~\ref{fig:stat_base_samples} shows the performance of these episodes in terms of return, return without $r_{\mathrm{bonus}}$, the number of timesteps the pendulum spent in the zenith-state, and the timestep in which the pendulum first reached this state.

Finding the most helpful samples to feed into the tree-learning algorithm is a nontrivial task. Given the episodes' length of $1000$ and the median time to first reach the zenith of $151$ timesteps, taking all samples would overrepresent a state in which the pendulum is in the upright position. DTs trained on these samples fail to swing the system to its state of maximum potential energy. It is therefore important, to limit the samples to the first $t_c$ timesteps of each episode. Extensive pre-studies lead to $t_c=350$ as a reasonable compromise, yielding good results. For the base samples, as well as for all following episodes in the iterative process, we therefore only include the first $350$ timesteps of each episode.

\subsection{Iterative DT training}
\begin{figure}
    \centering
    \includegraphics[width=\linewidth]{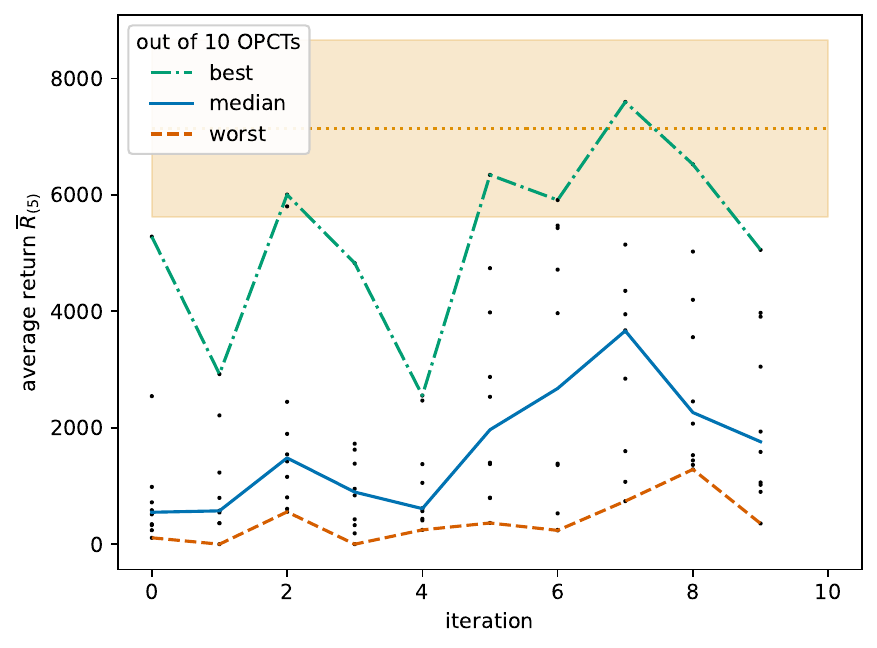}
    \caption{Performance of DTs evolving with iterations. Each dot represents the average return of one DT in $n_e=5$ evaluation episodes. With dashed red, solid blue, and dash-dotted green lines respectively, the worst, median, and best out of the $N_T=10$ DTs are connected. The orange dotted line and shaded area mark the DQN's return of $\overline{R} = 7138.83 \pm 1517.47$ in the $100$ episodes from which the $92$ episodes for the base samples were selected.}
    \label{fig:results}
\end{figure}

\begin{figure}
    \centering
    \includegraphics[width=\linewidth]{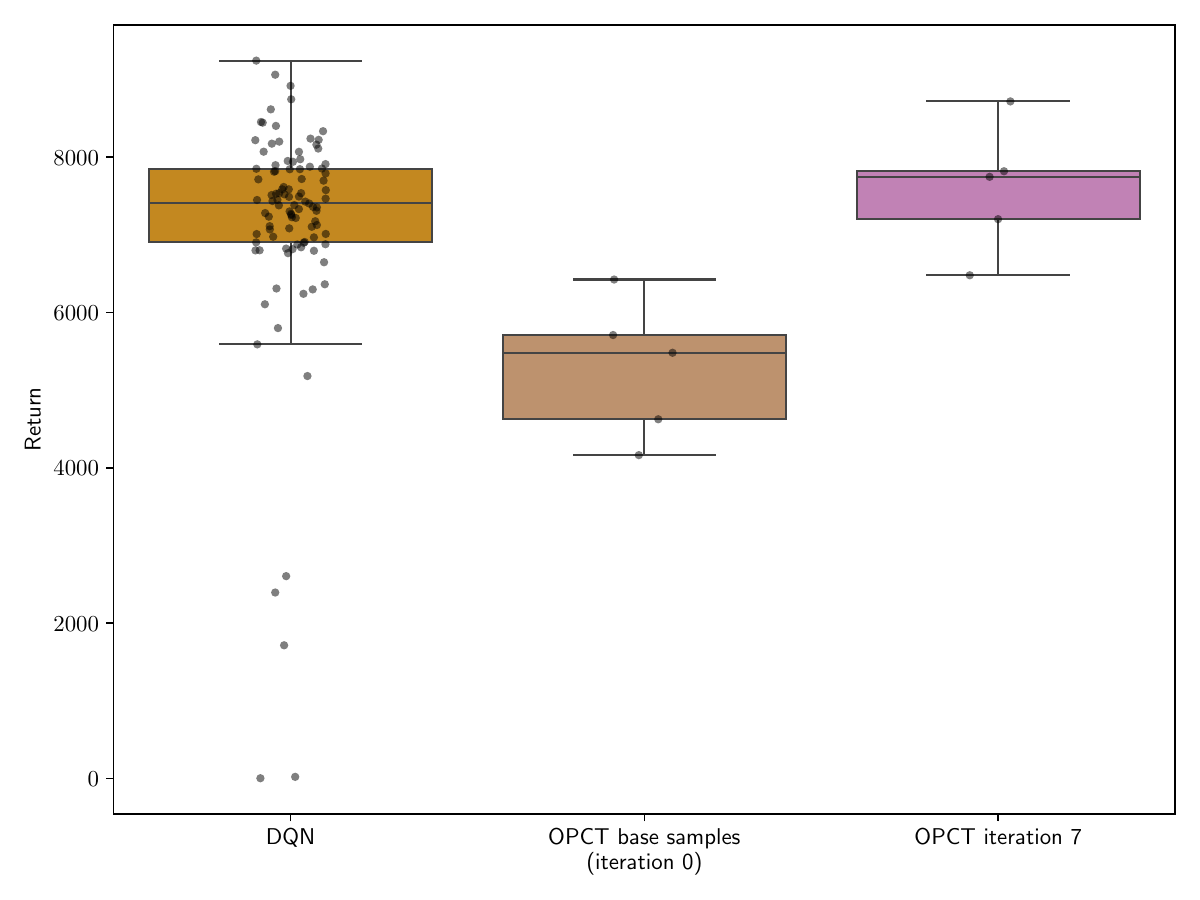}
    \caption{Comparison between the performance of the DQN, the DT trained on plain samples (iteration $0$), and the DT trained with the iterative algorithm. Shown are boxplots for the returns of the $100$ episodes of the DQN, and the returns of the $n_e=5$ episodes of the best DT in iteration $0$ and $7$ respectively. Each dot marks the return of a single episode. Also shown are the six outlier DQN episodes that were discarded when compiling the base samples.}
    \label{fig:comp}
\end{figure}

As expected, in iteration $0$, DTs trained only on the base samples exhibit a rather poor performance (as seen from the leftmost `iteration 0' results in Figure~\ref{fig:results} and the central boxplot of Figure~\ref{fig:comp}). The pendulum rarely reaches its zenith. 
In each following iteration, the first $t_c=350$ states of each of the $n_e=5$ evaluation episodes of the best-performing DT are labeled with the actions by the DQN and added to the previous samples. The resulting DTs are therefore trained on gradually larger datasets that cover more relevant regions of the state space. The fluctuations of the $N_T=10$ DTs trained in each iteration are rather large. However, an upward trend is observable in Figure~\ref{fig:results} after iteration $4$. After iteration $7$, the performance decreases again. This could be the consequence of overfitting. The best-performing DT (determined by the best average return in the $n_e=5$ evaluation episodes) is reached in iteration $7$. With its performance of $\overline{R} = 7594.87 \pm 826.85$, it is on par with the DQN used to label the states.

Figure~\ref{fig:comp} emphasized the effect of the iterative generation of samples by visualizing the performance of the DQN used as oracle, the DTs obtained from plain samples (iteration $0$), and the DTs obtained by the iterative algorithm (iteration $7$). While DTs trained on plain episodic samples do not reach the performance of the DQN agent, the iterative approach generates a sufficiently diverse dataset to successfully train DTs, able to match the performance of the DQN.

Videos of five episodes played by the best-performing DT are available on the Github repository.\footnote{\href{https://github.com/RaphaelEngelhardt/realworld_swingup}{https://github.com/RaphaelEngelhardt/realworld\_swingup}} Episode two shows an interesting behavior that is occasionally observed independently of whether the DT or the DQN is used as agent: If the agent manages to steer the pole in the upright position with no angular velocity left, the pole is balanced without any further actions, solely by the internal friction of the hinge.

\section{Discussion and Outlook}
We could show, that the iterative algorithm to train DTs is successfully applicable not only to simulated, but also to real-world robotic tasks. It should be noted, that this experiment primarily proves the feasibility; the results are subject to further improvement. A first remark goes toward the ratio of base samples and samples that are added at each iteration. In our experiment, the iterations start with $N_b=32200$ base samples ($92 \text{~episodes} \cdot 350 \text{~samples/episode}$), while each iteration only adds $1750$ samples ($5 \text{~episodes} \cdot 350 \text{~samples/episode}$). This ratio would require many iterations until the number of additional samples becomes significant w.r.t. the base samples. This problem is related to the time costs of playing evaluation episodes: Due to the fluctuations of the $N_T=10$ DTs generated in each iteration and within the $n_e=5$ evaluation episodes, these repetitions are needed. However, non-simulated episodes that are played in the real world take a substantial amount of time. 

It should be noted, that the resulting DT is not fully populated. While a binary DT of depth $d=10$ generally has $1023$ decision and $1024$ leaf nodes, our best-performing DT consists of $599$ decision nodes and $600$ leaves. The applied OPCT algorithm generates a distribution over the three possible actions at each leaf, the actual prediction is subsequently obtained via the $\argmax$ function. This allows for additional lossless pruning by collapsing branches whose leaves all yield the same result after applying $\argmax$ to the generally different distributions to a single leaf node containing the common prediction.\footnote{As an example, a decision node whose child nodes are leaves containing the distributions $[0.8, 0.1, 0.1]$ and $[0.7, 0.2, 0.1]$ respectively can simply be replaced by a leaf node containing the label `action 0'.} Doing so results in our best-performing DT comprising only $497$ decision and $498$ leaf nodes. This ultimately corresponds to a reduction of parameters of about $36\%$ compared to the DQN. 
More aggressive pruning could be done, for example by removing branches that are visited very seldom during episodes and therefore potentially only have a negligible effect on the overall performance.

In subsequent studies, the hyperparameters of the experiment should be optimized. This would allow for additional runs with decreasing depth $d$ of DTs. After this initial proof-of-concept, demonstrating the applicability of the iterative algorithm to real-world robotic tasks, we feel confident, that even simpler DTs could be found to solve the CPSU challenge.

\section*{Acknowledgment}
The authors express their gratitude to Prof. Tichelmann for allowing us to use the experimental setup in the Lab for Applied Artificial Intelligence at TH Köln -- University of Applied Sciences. 
We also appreciate the work previously done by Yendoukon Nayante to train the DQN that we used here as oracle.

\end{document}